\pdfoutput=1

\documentclass[11pt]{article}

\usepackage[]{acl2023}

\usepackage{times}
\usepackage{latexsym}
\usepackage{float}

\usepackage[T1]{fontenc}

\usepackage[utf8]{inputenc}

\usepackage{microtype}

\usepackage{amsmath,amsthm,mathtools}
\usepackage{microtype}
\usepackage[shortlabels]{enumitem}
\usepackage{soul}
\usepackage{makecell}
\usepackage{latexsym}
\usepackage{booktabs}
\usepackage{arydshln}
\DeclareMathOperator*{\argmax}{argmax}
\usepackage{color, colortbl}
\definecolor{myblue}{RGB}{51, 102, 153}
\definecolor{mylgreen}{RGB}{181, 234, 245}
\definecolor{mydblue}{RGB}{20, 136, 199}
\definecolor{myred}{RGB}{84, 19, 24}
\definecolor{myyellow}{RGB}{253, 253, 153}
\definecolor{mygreen}{RGB}{48, 128, 71}
\usepackage{multirow}
\usepackage{soul}
\usepackage{amssymb}
\usepackage{pifont}
\newcommand{\xmark}{{\color{myred}{\ding{55}}}\ }%
\newcommand{\xarrow}{{\color{myblue}{\ding{220}}}}
\title{Faithfulness Tests for Natural Language Explanations}
\author{
\parbox{\linewidth}{\centering
Pepa Atanasova$^{1}$, 
Oana-Maria Camburu$^{2}$,
Christina Lioma$^{1}$,
Thomas Lukasiewicz$^{3,4}$,
Jakob Grue Simonsen$^{1}$, 
Isabelle Augenstein$^{1}$}\\
\parbox{\linewidth}{\centering
  $^{1}$Department of Computer Science, University of Copenhagen, Denmark,
  $^{2}$University College London, UK, $^{3}$TU Wien, Austria, $^{4}$University of Oxford, UK \\
  \texttt{pepa@di.ku.dk}
  }
  }
\usepackage{bibentry}

\begin{document}
\maketitle
\begin{abstract}

Explanations of neural models aim to reveal a model's decision-making process for its predictions. However, recent work shows that current methods giving explanations such as saliency maps or counterfactuals can be misleading, as they are prone to present reasons that are unfaithful to the model's inner workings. This work explores the challenging question of evaluating the faithfulness of natural language explanations (NLEs). To this end, we present two tests. First, we propose a counterfactual input editor for inserting reasons that lead to counterfactual predictions but are not reflected by the NLEs. Second, we reconstruct inputs from the reasons stated in the generated NLEs and check how often they lead to the same predictions. Our tests can evaluate emerging NLE models, proving a fundamental tool in the development of faithful NLEs.


\end{abstract}
\section{Introduction}
\begin{table*}[t]
\fontsize{10}{10}\selectfont
\centering
\resizebox{1\textwidth}{!}{
\begin{tabular}{@{}>{\raggedright\arraybackslash}p{0.08\linewidth}p{0.65\linewidth}p{0.60\linewidth}@{}}
\toprule
\textbf{Test} & \textbf{Original Instance} & \textbf{Instance After Test Intervention}\\
\midrule
Counter\-fac\-tual test (\S\ref{sec:rq1}) & \textit{Premise:} Man in a black suit, white shirt and black bowtie playing an instrument with the rest of his symphony surrounding him. \newline \textit{Hypothesis:} A tall person in a suit. \newline \textit{Prediction:} neutral \newline \textit{NLE:} Not all men are tall. & \textit{Premise:} Man in a black suit, white shirt and black bowtie playing an instrument with the rest of his symphony surrounding him. \newline \textit{\xarrow Hypothesis:} A tall person in a {\color{myblue} blue} suit. \newline \textit{Prediction:} contradiction \newline \xmark  \textit{NLE:} \ul{A man is not a tall person.} \newline \color{myred}{\textit{Unfaithfulness cause:} inserted word \textbf{`blue'} $\notin$ NLE} but changed the prediction.\\

Input reconstruction test (\S\ref{sec:rq3}) & \textit{Premise:} Many people standing outside of a place talking to each other in front of a building that has a sign that says `HI-POINTE.' \newline \textit{Hypothesis:} The people are having a chat before going into the work building. \newline \textit{Prediction:} neutral \newline \textit{NLE:} \ul{Just because people are talking does not mean they are having a chat.} & \textit{\xarrow {Premise:}} {\color{myblue} People are talking.} \newline \textit{\xarrow {Hypothesis:}} {\color{myblue} They are having a chat.} \newline \xmark \textit{Prediction:} entailment \newline \textit{NLE:} People are talking is a rephrasing of they are having a chat. \newline \color{myred}{\textit{Unfaithfulness cause:} The reasons in the NLE for the original instance lead to a different prediction.}\\
\bottomrule
\end{tabular}}
\caption{Examples of unfaithful explanations detected with our tests for the task of NLI (see \S\ref{sec:method}). We apply the tests on an original instance (second column), which results in a new instance (third column). The parts of the input changed by the test are marked with \xarrow, and the intervention made by the test is in {\color{myblue} blue}. \xmark marks an NLE or a prediction that does not match the expectation, thus pointing to the \ul{underlined} NLE as being unfaithful. 
} 
\label{tab:examples}
\end{table*}

Explanations of neural models aim to uncover the reasons behind model predictions in order to provide evidence on whether the model is trustworthy. 
To this end, explanations have to be \textit{faithful}, i.e., reflect the decision-making process of the model, otherwise, they could be harmful 
\citep{hancox2020robustness}. 
However, recent studies show that explanations can often be unfaithful, 
covering flaws and biases of the model. \citet{adebayo2018sanity} show that certain widely deployed explainability approaches that provide saliency maps (with importance scores for each part of the input, e.g., words or super-pixels) can even be \emph{independent} of the training data or of the model parameters. 
Others 
also question the effectiveness and reliability of counterfactuals \cite{slack2021counterfactual}, concept activations, and training point ranking explanations \cite{adebayo2022post}.

In this work, we investigate the degree of faithfulness of natural language explanations 
(NLEs), 
which explain model predictions with free text. NLEs are not constrained to contain only input seg\-ments, thus they provide more expressive \citep{struggles} and usually more human-readable explanations than, e.g., saliency maps 
\citep{wiegreffe2021teach}. Evaluating the faithfulness of explanations is very challenging in general, as the ground-truth reasons used by a model for a prediction are usually unknown. Evaluating the faithfulness of NLEs is further complicated, as they often include words not present in the input. 
Thus, existing tests evaluating other types of explanations, e.g., saliency maps, cannot be directly applied to NLEs. 
As a stepping stone towards evaluating how faithful NLEs are, we design two tests. 
Our first test investigates whether NLE models are faithful to reasons for counterfactual predictions. We introduce 
a \textit{counterfactual input editor} that makes counterfactual interventions resulting in new instances on which the model prediction changes but the NLE does not reflect the intervention leading to the change. 
Our second test recon\-structs an input from the reasons stated in a generated NLE, and checks whether the new input leads to a different prediction. 
We apply our tests to four NLE models over three datasets. We aim for our tests to be an important tool 
to assess the faithfulness of existing and upcoming NLE models.\footnote{The code is available at \url{https://github.com/copenlu/nle_faithfulness}}

\section{The Faithfulness Tests} \label{sec:method}



Given a dataset $X {=} {(x_i, e_i, y_i)}$, 
with an input $x_i$, a gold NLE $e_i$, and a gold label $y_i \in L$, where $L$ is the set of all labels for $X$, a model $f$ is trained to produce an NLE and a task prediction for each input: $f(x_i)$ = ($\hat{e_i}$, $\hat{y_i}$). We also refer to $\hat{e_i}$ as $f(x_i)_{ex}$ and to $\hat{y_i}$ as $f(x_i)_p$.

\begin{table}[t!]
\fontsize{10}{10}\selectfont
\centering\resizebox{.4\textwidth}{!}{\begin{tabular}{llrr@{}r@{}}
\toprule
\textbf{Model} & \textbf{\%Counter} & \textbf{\thead{\%Counter \\ Unfaith}} & \textbf{\thead{\%Total \\ Unfaith}}\\
\midrule
\multicolumn{4}{c}{\textbf{e-SNLI}} \\
MT-Re-Rand & 38.85 & \textbf{60.39} & 23.46 \\
MT-Re-Edit & \textbf{56.70} & 46.12 & \textbf{26.15} \\
\textit{MT-Re-Rand+Edit} & \textit{64.98} & \textit{53.29} & \textit{34.63 }\\

ST-Re-Rand & 37.14 & \textbf{54.26} & 20.15 \\
ST-Re-Edit & \textbf{49.64} & 52.74 & \textbf{26.18} \\
\textit{ST-Re-Rand+Edit} & \textit{61.15} &\textit{58.27} & \textit{35.63}\\

MT-Ra-Rand & 37.17 & \textbf{54.93 }& 20.42 \\
MT-Ra-Edit & \textbf{55.04} & 41.34 & \textbf{22.75} \\
\textit{MT-Ra-Rand+Edit} & \textit{63.84} & \textit{48.63} & \textit{31.05} \\

ST-Ra-Rand & 35.21 & \textbf{57.82} & 20.36 \\
ST-Ra-Edit & \textbf{60.00} & 45.66 & \textbf{27.39 }\\
\textit{ST-Ra-Rand+Edit} & \textit{67.31} & \textit{55.03} &\underline{\textit{37.04}}\\

\multicolumn{4}{c}{\textbf{CoS-E}} \\
MT-Re-Rand & 44.89 & \textbf{83.18} &  37.34 \\
MT-Re-Edit & \textbf{50.00} & 77.23 &  \textbf{38.62} \\
\textit{MT-Re-Rand+Edit} & \textit{59.89} & \textit{85.26 }&  \textit{51.06} \\

ST-Re-Rand & 52.34 & 79.47 &  41.60 \\
ST-Re-Edit & \textbf{53.83} & \textbf{86.17} &  \textbf{46.38} \\
\textit{ST-Re-Rand+Edit} & \textit{67.45} & \textit{87.54} &  \underline{\textit{59.04}} \\

MT-Ra-Rand & 39.26 & \textbf{84.01} &  32.98 \\
MT-Ra-Edit & \textbf{50.00} & 78.72 & \textbf{ 39.36} \\
\textit{MT-Ra-Rand+Edit} & \textit{56.81} &\textit{ 85.58} &  \textit{48.62} \\

ST-Ra-Rand & 46.70 & \textbf{75.85} &  35.43 \\
ST-Ra-Edit & \textbf{52.02} & 75.05 &  \textbf{39.04 }\\
\textit{ST-Ra-Rand+Edit }& \textit{63.62} & \textit{81.77} &  \textit{52.02} \\

\multicolumn{4}{c}{\textbf{ComVE}} \\
MT-Re-Rand & 35.60 & \textbf{83.43} & 29.70 \\
MT-Re-Edit & \textbf{50.90} & 70.53 & \textbf{35.90} \\
\textit{MT-Re-Rand+Edit} &\textit{61.10} & \textit{78.89} &\textit{ 48.20} \\

ST-Re-Rand & 41.90 & 74.22 & 31.10 \\
ST-Re-Edit & \textbf{48.40} & \textbf{76.45} &\textbf{ 37.00} \\
\textit{ST-Re-Rand+Edit}& \textit{62.90} & \textit{77.42} & \textit{48.70} \\

MT-Ra-Rand & 33.70 & \textbf{75.67} & 25.50 \\
MT-Ra-Edit & \textbf{47.20 }& 66.53 & \textbf{31.40} \\
\textit{MT-Ra-Rand+Edit} & \textit{58.10 }& \textit{73.84} & \textit{42.90} \\

ST-Ra-Rand & 36.30 & \textbf{80.17} & 29.10 \\
ST-Ra-Edit &\textbf{49.50} & 79.80 & \textbf{39.50} \\
\textit{ST-Ra-Rand+Edit} & \textit{61.80 }& \textit{83.98} & \underline{\textit{51.90}} \\

\bottomrule
\end{tabular}}
\caption{Results for the \textbf{counterfactual test}. For each setup (Eq.~\ref{eq:setup}), we include the results of the random baseline (Rand), the counterfactual editor (Edit), and their union (Rand+Edit). The ``\% Coun\-ter'' column indicates the editor's success in finding inserts that change the model's prediction. 
``\% Counter Unfaith'' presents the percentage of instances where the inserted text was not found in the associated NLE among the instances where the prediction was changed. ``\% Total Unfaith'' presents the percentage of instances where the prediction was changed and the inserted text was not found in the associated NLE among all the instances in the test set. The highest rates of success in each pair of (Rand, Edit) tests are in bold. The highest total percentage of detected unfaithful NLEs for each dataset is underlined.}
\vspace*{-2ex}
\label{tab:counterfactual}
\end{table}

\smallskip \noindent
\textbf{2.1 \ The Counterfactual Test: Are NLE models faithful to reasons for counterfactual predictions?} \label{sec:rq1}
Studies in cognitive science show that humans usually seek counterfactuals by looking for factors that explain why event $\mathcal{A}$ occurred instead of $\mathcal{B}$ \cite{miller2019explanation}. 
Counterfactual explanations were proposed for ML models by making interventions either on the input \cite{wu-etal-2021-polyjuice,ross-etal-2021-explaining} or on the representation space \cite{jacovi-etal-2021-contrastive}. An intervention $h(x_i, y_i^C) = x_i'$ is produced over an input instance $x_i$ w.r.t.\ a target counterfactual label $y_i^C, y_i^C \neq \widehat{y_i}$, such that 
the model predicts the target label: $f(x_i') = \widehat{y_i'} = y_i^C$. 

For our test, we search for interventions that insert tokens into the input such that the model gives a different prediction, and we check whether the NLE reflects these tokens. 
%
%
Thus, we define an intervention $h(x_i, y_i^C) = x_i'$ that, for a given counterfactual label $y_i^C$, generates a set of words $W {=} \{w_j\}$ that, inserted into $x_i$, produces a new instance $x_i' = \{x_{i,1}, \dots x_{i,k}, W,$ $x_{i,k+1}, \dots x_{i,|x_i|} \}$ such that  $f(x_i')_p = y_i^C$. While one can insert each word in $W$ at a different position in $x_i$, here we define $W$ to be a \textit{contiguous} set of words, which is computationally less expensive. 
As $W$ is the counterfactual for the change in prediction, 
then at least one word from $W$ should be present in the NLE for the counterfactual prediction: 
\vspace*{-2ex}

{\begin{equation}~\label{eq:insertion} \small
\begin{split}
h(x_i, y_i^C) = x_i' \\
x_i' = \{x_{i,1}, \dots x_{i,k}, W, x_{i,k+1}, \dots x_{i,|x_i|} \} \\
f(h(x_i, y_i^C)) = f(x_i') = y_i^C \ne \widehat{y_i} = f(x_i) \\
\textrm{If } W \cap^s \widehat{e_i}' = \emptyset , \textrm{then $\widehat{e_i}'$ is unfaithful}, \\
\end{split}
\end{equation}}%
\noindent where the $s$ superscript indicates that the operator is used at the semantic level. 
Sample counterfactual interventions satisfying Eq.\,\ref{eq:insertion} are in Table~\ref{tab:examples}. More examples are in Tables~\ref{tab:random-examples} and \ref{tab:editor-examples} in the Appendix. 

To generate the input edits $W$, 
we propose an editor $h$ as a neural model 
%
and follow \citet{ross-etal-2021-explaining}. The authors generate input edits that change the model prediction to target 
predictions and refer to these edits as explanations. We note that besides the input edits, confounding factors could cause the change in prediction, e.g., the edits could make the model change its focus towards other parts of the input
and not base its decision on the edit itself. In this work, we presume that it is still important for the NLEs to point to the edits, as the model changed its prediction when the edit was inserted. This aligns with the literature on counterfactual explanations, where such edits are seen as explanations \cite{guidotti2022counterfactual}. We also hypothesize that confounding factors are rare, especially when insertions rather than deletions are performed. We leave such investigation for future work.   

During the training of $h$, we mask $n_1\%$ tokens in $x_i$, provide as an input to $h$ the label predicted by the model, i.e., $y_i^C = \widehat{y_i}$, and use the masked tokens to supervise the generation of the masked text (corresponding to $W$). 
During inference, we provide as target labels $y_i^C \in Y, y_i^C \neq \widehat{y_i}$, 
and we search over $n_2$ different positions to insert $n_3$ candidate tokens at each position at a time. 
The training objective is the cross-entropy loss for generating the inserts. 

We use as a metric of unfaithfulness the percentage of the instances in the test set for which $h$ finds counterfactual interventions that satisfy Eq.~\ref{eq:insertion}. 
To compute this automatically, we use $\cap^{s}$ at the syntactical level. As paraphrases of $W$ might appear in the NLEs, we manually verify a subset of NLEs. 
We leave the introduction of an automated evaluation for the semantic level for future work. 

Our metric is not a complete measure of the overall faithfulness of the NLEs, as (1) we only check whether the NLEs are faithful to the reasons for counterfactual predictions, and (2) it depends on the performance of $h$. But if $h$ does not succeed in finding a significant number of counterfactual reasons not reflected in the NLEs, it could be seen as evidence of the faithfulness of the model's NLEs.

\smallskip \noindent
\textbf{2.2 \ The Input Reconstruction Test: Are the reasons in an NLE sufficient to lead to the same prediction as the one for which the NLE was generated?} \label{sec:rq3}
Existing work points out that for an explanation to be faithful to the underlying model, the \textit{reasons $r_i$ in the explanation} should be \textit{sufficient} for the model to make the same prediction as on the original input \cite{yu-etal-2019-rethinking}:
{\begin{equation}\small
\begin{split}
    r_i = R(x_i, \widehat{e_i}) \\
     \textrm{If }f(r_i)_p \neq f(x_i)_p \textrm{, then }\widehat{e_i} \textrm{ is unfaithful},
\end{split}
\end{equation}}%
where $R$ is the function that builds a new input $r_i$ given $x_i$ and $\widehat{e_i}$.
Sufficiency has been employed to evaluate saliency explanations, where the direct mapping between tokens and saliency scores allows $r_i$ to be easily constructed (by preserving only the top-N most salient tokens) \cite{deyoung-etal-2020-eraser,atanasova-etal-2020-diagnostic}. 
For NLEs, which lack such direct mapping, designing an automated extraction $R$ of the reasons in $\widehat{e_i}$ is challenging. 

Here, we propose automated agents $R$s that are task-dependent. 
%
We build $R$s for 
e-SNLI \cite{NIPS2018_8163} and ComVE \cite{wang-etal-2020-semeval}, due to the structure of the NLEs and the nature of these datasets. However, we could not construct an $R$ for CoS-E \citep{rajani-etal-2019-explain}.  
For e-SNLI, a large number of NLEs follow certain templates.
\citet{camburu-etal-2020-make} provide a list of templates covering 97.4\% of the NLEs in the training set. For example, ``<X> is the same as <Y>'' is an NLE template for entailment. Thus, many of the generated NLEs also follow these templates. In our test,  we simply use <X> and <Y> from the templates as the reconstructed pair of premise and hypothesis, respectively. We keep only those <X> and <Y> that are sentences containing at least one subject and at least one verb. If the NLE for the original input was faithful, then we expect the prediction for the reconstructed input to be the same as for the original.  

Given two sentences, the ComVE task is to pick the one that contradicts common sense. 
If the generated NLE is faithful, replacing the correct sentence with the NLE should lead to the same prediction. 
\section{Experiments} \label{sec:experiments}
\begin{table}[t!]
\fontsize{10}{10}\selectfont
\centering\resizebox{.40\textwidth}{!}{\begin{tabular}{@{}llrr@{}}
\toprule
& \textbf{Model} & \textbf{\% Reconst} & \textbf{\% Total Unfaith}\\
\midrule
{\textbf{e-SNLI}} & MT-Re & 39.49 & 7.7 \\
& ST-Re & 39.99 & \textbf{9.7} \\
& MT-Ra & 44.87 & 7.8 \\
& ST-Ra & 43.32 & 9.3 \\
 {\textbf{ComVE}} & MT-Re & 100 & 36.9  \\
& ST-Re & 100 & 22.7  \\
& MT-Ra & 100 & \textbf{40.3} \\
& ST-Ra & 100 & 28.5  \\
\bottomrule
\end{tabular}}
\caption{Results for the \textbf{input reconstruction test}. ``\%~Reconst'' shows the percentage of instances for which we managed to form a reconstructed input. ``\%~Total Unfaith'' shows the total percentage of unfaithful NLEs found among all instances in the test set of each dataset. The highest detected percentage of unfaithful NLEs for each dataset is in bold.} 
\label{tab:consistency}\vspace*{-2.5ex}
\end{table}
Following \citet{hase-etal-2020-leakage}, we experiment with four setups for NLE models, 
which can be grouped by whether the prediction and NLE generation 
are trained with a multi-task objective using a joint model (MT) or with single-task objectives using separate models (ST). They can also be grouped by whether they generate NLEs conditioned on the predicted label (rationalizing models (Ra)), or not conditioned on it (reasoning models (Re)). The general notation $f(x_i) = (\widehat{e_i}, \widehat{y_i})$ used in \S\ref{sec:method} includes all four setups: 
\vspace*{-2.5ex}

{\begin{equation}\small
\begin{split}
    \text{\bf{MT-Re:}} f_{p,ex}(x_i) = (\widehat{e_i}, \widehat{y_i}) \\
    \text{\bf{MT-Ra:}} f_{p,ex}(x_i) = (\widehat{e_{i|\widehat{y_i}\footnotemark}}, \widehat{y_i})\\
    \text{\bf{ST-Re:}} f_{ex}(x_i) = \widehat{e_i} ; 
    f_p(x_i, \widehat{e_i}) = \widehat{y_i} \\
    \text{\bf{ST-Ra:}} f_{ex}(x_i, y_{j}) = \widehat{e_{i,j}}; f_p(x_i, \widehat{e_i}) = \widehat{y_j}\ \\ 
    j = \argmax\nolimits_{j \in [1,...,|L|]}(f_{p}(x_i, \widehat{e_{i,j}})),
\end{split}
\label{eq:setup}
\end{equation}}%
where $f_{p,ex}$ is a joint model for task prediction and NLE generation, $f_{p}$ is a model only for task prediction, and $f_{ex}$ is a model only for NLE generation. 
The ST-Ra setup produces one NLE $e_{i,j}$ for each $y_j \in L$. Given $\widehat{e_{i,j}}$ and $x_i$, $f_{p}$ predicts the probability of the corresponding label $y_j$ and selects as $\widehat{y_i}$ the label with the highest probability.

\footnotetext{During training, the gold label is used.}

For both $f$ and the editor $h$, we employ the pre-trained T5-base model \cite{raffel2020exploring}. 
The editor uses task-specific prefixes for insertion and NLE generation. We train both $f$ and $h$ for 20 epochs, evaluate them on the validation set at each epoch, and select the checkpoints with the highest success rate (see \S\ref{sec:rq1}). We use a learning rate of 1e-4 with the Adam optimizer \cite{kingma2014adam}. 
For the editor, during training, we mask $n_1$ consecutive tokens with one mask token, where $n_1$ is chosen at random in $ [1, 3]$. 
During inference, we generate candidate insertions for $n_2 = 4$ random positions, with $n_3 = 4$ candidates for each position at a time. 
The hyper-parameters are chosen with a grid search over the validation set.\footnote{When $n_2$ and $n_3$ are increased, a higher number of insertions are generated, which in turn could result in a higher percentage of unfaithful NLEs. However, increasing these parameters also leads to higher computational demands. Future research could explore strategies for efficiently searching the space of insertion candidates.} 
For the manual evaluation
, an author annotated the first 100 test instances for each model (800 in total). The manual evaluation has been designed in accordance with related work \cite{NIPS2018_8163}, which also evaluated 100 instances per model. We found that no instances were using paraphrases. Hence, in our work, the automatic metric can be trusted. 

\noindent 
\textbf{Baseline.}
For the counterfactual test, we incorporate a random baseline as a comparison. Specifically, we insert a random adjective before a noun or a random adverb before a verb. We randomly select $n_2 = 4$ positions where we insert the said words, and, for each position at a time, we consider $n_3 = 4$ random candidate words. The candidates are single words randomly chosen from the complete list of adjectives or adverbs available in WordNet \cite{fellbaum2010wordnet}. We identify the nouns and verbs in the text with 
spaCy \cite{spacy2}.

\noindent  
\textbf{Datasets.}
We use three popular datasets with NLEs: e-SNLI \cite{NIPS2018_8163}, CoS-E \cite{rajani-etal-2019-explain}, and ComVE \cite{wang-etal-2020-semeval}. 
e-SNLI contains NLEs for SNLI \cite{bowman-etal-2015-large}, where, given a premise and a hypothesis, one has to predict whether they are in a relationship of \textit{entailment} (the premise entails the hypothesis), \textit{contradiction} (the hypothesis contradicts the premise), or \textit{neutral} (neither entailment nor contradiction hold).
CoS-E contains NLEs for commonsense question answering, where given a question, one has to pick the correct answer out of three given options. 
ComVE contains NLEs for commonsense reasoning, where given two sentences, one has to pick the one that violates common sense.

\smallskip\noindent 
\textbf{3.1 \ Results}

\noindent\textbf{Counterfactual Test.}
Table \ref{tab:counterfactual} shows the results of our counterfactual test. 
First, we observe that when the random baseline finds words that change the prediction of the model, the words are more often not found in the corresponding NLE compared to the counterfactual editor (\% Counter Unfaith). We conjecture that this is because the randomly selected words are rare for the dataset compared to the words that the editor learns to insert. Second, the counterfactual editor is better at finding words that lead to a change in the model's prediction, which in turn results in a higher percentage of unfaithful instances in general (\% Total Unfaith). 
We also observe that the insertions $W$ lead to counterfactual predictions for up to 56.70\% of the instances (for MT-Re-Edit on e-SNLI).  
For up to 46.38\% of the instances (for ST-Re-Edit on CoS-E), the editor is able to find an insertion for which the counterfactual NLE is unfaithful. Table \ref{tab:examples}, row 1, presents one such example. More examples for the random baseline can be found in Table \ref{tab:random-examples}, and for the counterfactual editor in Table \ref{tab:editor-examples}. Finally, the union of the counterfactual interventions discovered by the random baseline and the editor, we observe total percentages of up to 59.04\% unfaithfulness to the counterfactual. 

We see that for all datasets and models, the total percentages of unfaithfulness to counterfactual are high, between 37.04\% (for MT-Ra-Rand+Edit on e-SNLI) and 59.04\% (ST-Re-Rand+Edit for CoS-E). 
We re-emphasize that this should not be interpreted as an overall estimate of unfaithfulness, as our test is not complete (see \S\ref{sec:rq1}). 

\noindent  
\textbf{The Input Reconstruction Test.}
Table \ref{tab:consistency} shows the~results of the input reconstruction test. 
We were able to reconstruct inputs for up to 4487 out of the 10K test instances in e-SNLI, and for all test instances in ComVE. There are, again, a substantial number of unfaithful NLEs: 
up to 14\% for e-SNLI, and up to 40\% for ComVE. An example is in Table \ref{tab:examples}, row 2. More examples can be found in Table \ref{tab:cosistency:examples}. We also notice that this test identified considerably more unfaithful NLEs for ComVE than for e-SNLI, while for our first test, the gap was not as pronounced. 
This shows the utility of developing diverse faithfulness tests. 

Finally, all four types of models had similar faithfulness results\footnote{Task accuracy and NLE quality are given in Table \ref{tab:performance}.} on all datasets and tests, with no consistent ranking among them. This opposes the intuition that some configurations may be more faithful than others, e.g., \citet{NIPS2018_8163} hypothesized that ST-Re may be more faithful than MT-Re, which is the case in most but not all of the cases, e.g., on CoS-E the editorial finds more unfaithfulness for ST-Re (44.04\%) than for MT-Re (42.76 \%). We also observe that Re models tend to be less faithful than Ra models in most cases.  

\section{Related Work}\vspace*{-0.75ex}
\noindent\textbf{Tests for Saliency Maps.}
The faithfulness and, more generally, the utility of explanations were predominantly explored for saliency maps. Comprehensiveness and sufficiency \cite{deyoung-etal-2020-eraser} were proposed for evaluating the faithfulness of existing saliency maps. They measure the decrease in a model's performance when only the most or the least important tokens are removed from the input. \citet{madsen-etal-2022-evaluating} propose another faithfulness metric for saliency maps, ROAR, obtained by masking allegedly important tokens and then retraining the model. 
In addition, \citet{yin-etal-2022-sensitivity} and \citet{hsieh2021evaluations} evaluate saliency maps through adversarial input manipulations presuming that model predictions should be more sensitive to manipulations of the more important input regions as per the saliency map. \citet{chan-etal-2022-comparative} provide a comparative study of faithfulness measures for saliency maps. Further faithfulness testing for saliency maps was introduced by \citet{verifyExplainers}. 
%
Existing studies also pointed out that saliency maps can be manipulated to hide a classifier's biases towards dataset properties such as gender and race \cite{NEURIPS2019_bb836c01,10.1145/3375627.3375830,anders2020fairwashing}. While diagnostic methods for saliency maps rely on the one-to-one correspondence between the saliency scores and the regions of the input, this correspondence is not present for NLEs, where text not in the input can be included. Thus, diagnostic methods for saliency maps are not directly applicable to NLEs. To this end, we propose diagnostic tests that can be used to evaluate NLE model faithfulness.

\noindent\textbf{Tests for NLEs.}
Existing work 
often only looks at the plausibility of the NLEs \citep{rajani-etal-2019-explain, kayser2021vil, marasovic2022, wt5, mimicnle, oodNLEs}. In addition, \citet{sun-etal-2022-investigating} investigated whether the additional context available in human- and model-generated NLEs can benefit model prediction as they benefit human users. Differently, \citet{hase-etal-2020-leakage} proposed to measure the utility of NLEs in terms of how well an observer can simulate a model's output given the generated NLE. The observer could be an agent \citep{chan2022frame} or a human \cite{info13100500,atanasova-etal-2020-generating-fact}. 
%
The only work we are aware of that introduces sanity tests for the faithfulness of NLEs is that of \citet{wiegreffe-etal-2021-measuring}, who suggest that an association between labels and NLEs is necessary for faithful NLEs and propose two pass/fail tests: 
(1) whether the predicted label and generated NLE are similarly robust to noise, (2) whether task prediction and NLE generation share the most important input tokens for each. \citet{pmlr-v162-majumder22a} use these tests as a sanity check for the faithfulness of their model.  
Our tests are complementary and offer quantitative metrics.


\section{Summary and Outlook} 
In this work, we introduced two tests to evaluate the faithfulness of NLE models. 
We find that all four high-level setups of NLE 
models are prone to generate unfaithful NLEs, reinforcing the need for proof of faithfulness. 
Our tests can be used to ensure the faithfulness of emerging NLE models and inspire the community to design complementary faithfulness tests. 

\section*{Limitations} 
While our tests are an important stepping stone for evaluating the faithfulness of NLEs, they are not comprehensive. Hence, a model that would perform perfectly on our tests may still generate unfaithful NLEs. 

Our first test inspects whether NLE models are faithful to reasons for counterfactual predictions. It is important to highlight that NLEs may not comprehensively capture all the underlying reasons for a model's prediction. Thus, an NLE that fails to accurately represent the reasons for counterfactual predictions may still offer faithful explanations by reflecting other relevant factors contributing to the predictions. Additionally, both the random baseline and the counterfactual editor can generate insertions that result in text lacking semantic coherence. To address this limitation, future research can explore methods to generate insertion candidates that are both semantically coherent and reveal unfaithful NLEs.

Our second test uses heuristics that are task-dependent and may not be applicable to any task. The reconstruction functions $R$s proposed in this work are based on hand-crafted rules for the e-SNLI and ComVE datasets. However, due to the nature of the CoS-E NLEs, rule-based input reconstructions were not possible for this dataset. To address this limitation, future research could investigate automated reconstruction functions that utilize machine learning models. These models would be trained to generate reconstructed inputs based on the generated NLEs, where a small number of annotations would be provided as training instances. For example, for CoS-E, one such training annotation could be: \textit{Original Question:} After getting drunk people couldn’t understand him, it was because of his what? \textit{Choices:} lower standards, slurred speech, or falling down. \textit{Answer:} slurred speech. \textit{NLE: }People who are drunk have difficulty speaking. $\rightarrow$ \textit{Reconstructed Question:} What do drunk people have difficulty with? \textit{Reconstructed Choices:} lower standards, speaking, or falling down. This approach would enable the development of machine learning models capable of generating reconstructed inputs for various datasets.

\section*{Acknowledgements}
$\begin{array}{l}\includegraphics[width=1cm]{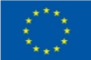} \end{array}$ The research documented in this paper has received funding from the European Union's Horizon 2020 research and innovation programme under the Marie Sk\l{}odowska-Curie grant agreement No 801199. Isabelle Augenstein's research is further partially funded by a DFF Sapere Aude research leader grant under grant agreement No 0171-00034B, as well as by the Pioneer Centre for AI, DNRF grant number P1.
Thomas Lukasiewicz was supported by the Alan Turing Institute under the UK EPSRC grant EP/N510129/1, the AXA Research Fund, and the EU TAILOR grant 952215. 
Oana-Maria Camburu was supported by a UK \mbox{Leverhulme} Early Career Fellowship.
Christina Lioma's research is partially funded by the Villum and Velux Foundations Algorithms, Data and Democracy (ADD) grant.


\bibliography{custom,anthology}

\begin{thebibliography}{43}
\expandafter\ifx\csname natexlab\endcsname\relax\def\natexlab#1{#1}\fi

\bibitem[{Adebayo et~al.(2018)Adebayo, Gilmer, Muelly, Goodfellow, Hardt, and
  Kim}]{adebayo2018sanity}
Julius Adebayo, Justin Gilmer, Michael Muelly, Ian Goodfellow, Moritz Hardt,
  and Been Kim. 2018.
\newblock Sanity checks for saliency maps.
\newblock \emph{Advances in Neural Information Processing Systems}, 31.

\bibitem[{Adebayo et~al.(2022)Adebayo, Muelly, Abelson, and
  Kim}]{adebayo2022post}
Julius Adebayo, Michael Muelly, Harold Abelson, and Been Kim. 2022.
\newblock \href {https://openreview.net/forum?id=xNOVfCCvDpM} {Post hoc
  explanations may be ineffective for detecting unknown spurious correlation}.
\newblock In \emph{International Conference on Learning Representations}.

\bibitem[{Anders et~al.(2020)Anders, Pasliev, Dombrowski, M{\"u}ller, and
  Kessel}]{anders2020fairwashing}
Christopher Anders, Plamen Pasliev, Ann-Kathrin Dombrowski, Klaus-Robert
  M{\"u}ller, and Pan Kessel. 2020.
\newblock Fairwashing explanations with off-manifold detergent.
\newblock In \emph{International Conference on Machine Learning}, pages
  314--323. PMLR.

\bibitem[{Atanasova et~al.(2020{\natexlab{a}})Atanasova, Simonsen, Lioma, and
  Augenstein}]{atanasova-etal-2020-diagnostic}
Pepa Atanasova, Jakob~Grue Simonsen, Christina Lioma, and Isabelle Augenstein.
  2020{\natexlab{a}}.
\newblock \href {https://doi.org/10.18653/v1/2020.emnlp-main.263} {A diagnostic
  study of explainability techniques for text classification}.
\newblock In \emph{Proceedings of the 2020 Conference on Empirical Methods in
  Natural Language Processing (EMNLP)}, pages 3256--3274, Online. Association
  for Computational Linguistics.

\bibitem[{Atanasova et~al.(2020{\natexlab{b}})Atanasova, Simonsen, Lioma, and
  Augenstein}]{atanasova-etal-2020-generating-fact}
Pepa Atanasova, Jakob~Grue Simonsen, Christina Lioma, and Isabelle Augenstein.
  2020{\natexlab{b}}.
\newblock \href {https://doi.org/10.18653/v1/2020.acl-main.656} {Generating
  fact checking explanations}.
\newblock In \emph{Proceedings of the 58th Annual Meeting of the Association
  for Computational Linguistics}, pages 7352--7364, Online. Association for
  Computational Linguistics.

\bibitem[{Bowman et~al.(2015)Bowman, Angeli, Potts, and
  Manning}]{bowman-etal-2015-large}
Samuel~R. Bowman, Gabor Angeli, Christopher Potts, and Christopher~D. Manning.
  2015.
\newblock \href {https://doi.org/10.18653/v1/D15-1075} {A large annotated
  corpus for learning natural language inference}.
\newblock In \emph{Proceedings of the 2015 Conference on Empirical Methods in
  Natural Language Processing}, pages 632--642, Lisbon, Portugal. Association
  for Computational Linguistics.

\bibitem[{Camburu et~al.(2019)Camburu, Giunchiglia, Foerster, Lukasiewicz, and
  Blunsom}]{verifyExplainers}
Oana-Maria Camburu, Eleonora Giunchiglia, Jakob Foerster, Thomas Lukasiewicz,
  and Phil Blunsom. 2019.
\newblock {Can I Trust the Explainer? {V}erifying Post-hoc Explanatory
  Methods}.
\newblock In \emph{NeurIPS 2019 Workshop Safety and Robustness in Decision
  Making}.

\bibitem[{Camburu et~al.(2021)Camburu, Giunchiglia, Foerster, Lukasiewicz, and
  Blunsom}]{struggles}
Oana-Maria Camburu, Eleonora Giunchiglia, Jakob Foerster, Thomas Lukasiewicz,
  and Phil Blunsom. 2021.
\newblock The struggles of feature-based explanations: Shapley values vs.
  minimal sufficient subsets.
\newblock In \emph{AAAI 2021 Workshop on Explainable Agency in Artificial
  Intelligence}.

\bibitem[{Camburu et~al.(2018)Camburu, Rockt\"{a}schel, Lukasiewicz, and
  Blunsom}]{NIPS2018_8163}
Oana-Maria Camburu, Tim Rockt\"{a}schel, Thomas Lukasiewicz, and Phil Blunsom.
  2018.
\newblock \href
  {http://papers.nips.cc/paper/8163-e-snli-natural-language-inference-with-natural-language-explanations.pdf}
  {{e-SNLI: Natural Language Inference with Natural Language Explanations}}.
\newblock In S.~Bengio, H.~Wallach, H.~Larochelle, K.~Grauman, N.~Cesa-Bianchi,
  and R.~Garnett, editors, \emph{Advances in Neural Information Processing
  Systems 31}, pages 9539--9549. Curran Associates, Inc.

\bibitem[{Camburu et~al.(2020)Camburu, Shillingford, Minervini, Lukasiewicz,
  and Blunsom}]{camburu-etal-2020-make}
Oana-Maria Camburu, Brendan Shillingford, Pasquale Minervini, Thomas
  Lukasiewicz, and Phil Blunsom. 2020.
\newblock \href {https://doi.org/10.18653/v1/2020.acl-main.382} {Make up your
  mind! adversarial generation of inconsistent natural language explanations}.
\newblock In \emph{Proceedings of the 58th Annual Meeting of the Association
  for Computational Linguistics}, pages 4157--4165, Online. Association for
  Computational Linguistics.

\bibitem[{Chan et~al.(2022{\natexlab{a}})Chan, Nie, Tan, Peng, Firooz, Sanjabi,
  and Ren}]{chan2022frame}
Aaron Chan, Shaoliang Nie, Liang Tan, Xiaochang Peng, Hamed Firooz, Maziar
  Sanjabi, and Xiang Ren. 2022{\natexlab{a}}.
\newblock Frame: Evaluating simulatability metrics for free-text rationales.
\newblock \emph{arXiv preprint arXiv:2207.00779}.

\bibitem[{Chan et~al.(2022{\natexlab{b}})Chan, Kong, and
  Guanqing}]{chan-etal-2022-comparative}
Chun~Sik Chan, Huanqi Kong, and Liang Guanqing. 2022{\natexlab{b}}.
\newblock \href {https://doi.org/10.18653/v1/2022.acl-long.345} {A comparative
  study of faithfulness metrics for model interpretability methods}.
\newblock In \emph{Proceedings of the 60th Annual Meeting of the Association
  for Computational Linguistics (Volume 1: Long Papers)}, pages 5029--5038,
  Dublin, Ireland. Association for Computational Linguistics.

\bibitem[{DeYoung et~al.(2020)DeYoung, Jain, Rajani, Lehman, Xiong, Socher, and
  Wallace}]{deyoung-etal-2020-eraser}
Jay DeYoung, Sarthak Jain, Nazneen~Fatema Rajani, Eric Lehman, Caiming Xiong,
  Richard Socher, and Byron~C. Wallace. 2020.
\newblock \href {https://doi.org/10.18653/v1/2020.acl-main.408} {{ERASER}: {A}
  benchmark to evaluate rationalized {NLP} models}.
\newblock In \emph{Proceedings of the 58th Annual Meeting of the Association
  for Computational Linguistics}, pages 4443--4458, Online. Association for
  Computational Linguistics.

\bibitem[{Dombrowski et~al.(2019)Dombrowski, Alber, Anders, Ackermann,
  M\"{u}ller, and Kessel}]{NEURIPS2019_bb836c01}
Ann-Kathrin Dombrowski, Maximillian Alber, Christopher Anders, Marcel
  Ackermann, Klaus-Robert M\"{u}ller, and Pan Kessel. 2019.
\newblock \href
  {https://proceedings.neurips.cc/paper/2019/file/bb836c01cdc9120a9c984c525e4b1a4a-Paper.pdf}
  {Explanations can be manipulated and geometry is to blame}.
\newblock In \emph{Advances in Neural Information Processing Systems},
  volume~32. Curran Associates, Inc.

\bibitem[{Fellbaum(2010)}]{fellbaum2010wordnet}
Christiane Fellbaum. 2010.
\newblock Wordnet.
\newblock In \emph{Theory and Applications of Ontology: Computer Applications},
  pages 231--243. Springer.

\bibitem[{Guidotti(2022)}]{guidotti2022counterfactual}
Riccardo Guidotti. 2022.
\newblock Counterfactual explanations and how to find them: literature review
  and benchmarking.
\newblock \emph{Data Mining and Knowledge Discovery}, pages 1--55.

\bibitem[{Hancox-Li(2020)}]{hancox2020robustness}
Leif Hancox-Li. 2020.
\newblock Robustness in machine learning explanations: does it matter?
\newblock In \emph{{Proceedings of the 2020 Conference on Fairness,
  Accountability, and Transparency}}, pages 640--647.

\bibitem[{Hase et~al.(2020)Hase, Zhang, Xie, and
  Bansal}]{hase-etal-2020-leakage}
Peter Hase, Shiyue Zhang, Harry Xie, and Mohit Bansal. 2020.
\newblock \href {https://doi.org/10.18653/v1/2020.findings-emnlp.390}
  {Leakage-adjusted simulatability: Can models generate non-trivial
  explanations of their behavior in natural language?}
\newblock In \emph{Findings of the Association for Computational Linguistics:
  EMNLP 2020}, pages 4351--4367, Online. Association for Computational
  Linguistics.

\bibitem[{Honnibal et~al.(2020)Honnibal, Montani, Van~Landeghem, and
  Boyd}]{spacy2}
Matthew Honnibal, Ines Montani, Sofie Van~Landeghem, and Adriane Boyd. 2020.
\newblock {spaCy: Industrial-strength Natural Language Processing in Python}.

\bibitem[{Hsieh et~al.(2021)Hsieh, Yeh, Liu, Ravikumar, Kim, Kumar, and
  Hsieh}]{hsieh2021evaluations}
Cheng-Yu Hsieh, Chih-Kuan Yeh, Xuanqing Liu, Pradeep~Kumar Ravikumar, Seungyeon
  Kim, Sanjiv Kumar, and Cho-Jui Hsieh. 2021.
\newblock \href {https://openreview.net/forum?id=4dXmpCDGNp7} {Evaluations and
  methods for explanation through robustness analysis}.
\newblock In \emph{International Conference on Learning Representations}.

\bibitem[{Jacovi et~al.(2021)Jacovi, Swayamdipta, Ravfogel, Elazar, Choi, and
  Goldberg}]{jacovi-etal-2021-contrastive}
Alon Jacovi, Swabha Swayamdipta, Shauli Ravfogel, Yanai Elazar, Yejin Choi, and
  Yoav Goldberg. 2021.
\newblock \href {https://aclanthology.org/2021.emnlp-main.120} {Contrastive
  explanations for model interpretability}.
\newblock In \emph{Proceedings of the 2021 Conference on Empirical Methods in
  Natural Language Processing}, pages 1597--1611, Online and Punta Cana,
  Dominican Republic. Association for Computational Linguistics.

\bibitem[{Jolly et~al.(2022)Jolly, Atanasova, and Augenstein}]{info13100500}
Shailza Jolly, Pepa Atanasova, and Isabelle Augenstein. 2022.
\newblock \href {https://doi.org/10.3390/info13100500} {{Generating Fluent Fact
  Checking Explanations with Unsupervised Post-Editing}}.
\newblock \emph{Information}, 13(10).

\bibitem[{Kayser et~al.(2021)Kayser, Camburu, Salewski, Emde, Do, Akata, and
  Lukasiewicz}]{kayser2021vil}
Maxime Kayser, Oana-Maria Camburu, Leonard Salewski, Cornelius Emde, Virginie
  Do, Zeynep Akata, and Thomas Lukasiewicz. 2021.
\newblock {e-ViL: A} dataset and benchmark for natural language explanations in
  vision-language tasks.
\newblock In \emph{Proceedings of the IEEE/CVF International Conference on
  Computer Vision}, pages 1244--1254.

\bibitem[{Kayser et~al.(2022)Kayser, Emde, Camburu, Parsons, Papiez, and
  Lukasiewicz}]{mimicnle}
Maxime Kayser, Cornelius Emde, Oana-Maria Camburu, Guy Parsons, Bartlomiej
  Papiez, and Thomas Lukasiewicz. 2022.
\newblock Explaining chest x-ray pathologies in natural language.
\newblock In \emph{Medical Image Computing and Computer Assisted Intervention
  -- MICCAI 2022}, pages 701--713, Cham. Springer Nature Switzerland.

\bibitem[{Kingma and Ba(2014)}]{kingma2014adam}
Diederik~P Kingma and Jimmy Ba. 2014.
\newblock Adam: A method for stochastic optimization.
\newblock \emph{arXiv preprint arXiv:1412.6980}.

\bibitem[{Madsen et~al.(2022)Madsen, Meade, Adlakha, and
  Reddy}]{madsen-etal-2022-evaluating}
Andreas Madsen, Nicholas Meade, Vaibhav Adlakha, and Siva Reddy. 2022.
\newblock \href {https://aclanthology.org/2022.findings-emnlp.125} {{Evaluating
  the Faithfulness of Importance Measures in {NLP} by Recursively Masking
  Allegedly Important Tokens and Retraining}}.
\newblock In \emph{Findings of the Association for Computational Linguistics:
  EMNLP 2022}, pages 1731--1751, Abu Dhabi, United Arab Emirates. Association
  for Computational Linguistics.

\bibitem[{Majumder et~al.(2022)Majumder, Camburu, Lukasiewicz, and
  Mcauley}]{pmlr-v162-majumder22a}
Bodhisattwa~Prasad Majumder, Oana Camburu, Thomas Lukasiewicz, and Julian
  Mcauley. 2022.
\newblock \href {https://proceedings.mlr.press/v162/majumder22a.html}
  {Knowledge-grounded self-rationalization via extractive and natural language
  explanations}.
\newblock In \emph{Proceedings of the 39th International Conference on Machine
  Learning}, volume 162 of \emph{Proceedings of Machine Learning Research},
  pages 14786--14801. PMLR.

\bibitem[{Marasović et~al.(2022)Marasović, Beltagy, Downey, and
  Peters}]{marasovic2022}
Ana Marasović, Iz~Beltagy, Doug Downey, and Matthew~E. Peters. 2022.
\newblock Few-shot self-rationalization with natural language prompts.
\newblock \emph{Findings of NAACL}.

\bibitem[{Miller(2019)}]{miller2019explanation}
Tim Miller. 2019.
\newblock Explanation in artificial intelligence: Insights from the social
  sciences.
\newblock \emph{Artificial intelligence}, 267:1--38.

\bibitem[{Narang et~al.(2020)Narang, Raffel, Lee, Roberts, Fiedel, and
  Malkan}]{wt5}
Sharan Narang, Colin Raffel, Katherine Lee, Adam Roberts, Noah Fiedel, and
  Karishma Malkan. 2020.
\newblock \href {http://arxiv.org/abs/2004.14546} {{WT5}?! training
  text-to-text models to explain their predictions}.

\bibitem[{Raffel et~al.(2020)Raffel, Shazeer, Roberts, Lee, Narang, Matena,
  Zhou, Li, Liu et~al.}]{raffel2020exploring}
Colin Raffel, Noam Shazeer, Adam Roberts, Katherine Lee, Sharan Narang, Michael
  Matena, Yanqi Zhou, Wei Li, Peter~J Liu, et~al. 2020.
\newblock Exploring the limits of transfer learning with a unified text-to-text
  transformer.
\newblock \emph{J. Mach. Learn. Res.}, 21(140):1--67.

\bibitem[{Rajani et~al.(2019)Rajani, McCann, Xiong, and
  Socher}]{rajani-etal-2019-explain}
Nazneen~Fatema Rajani, Bryan McCann, Caiming Xiong, and Richard Socher. 2019.
\newblock \href {https://doi.org/10.18653/v1/P19-1487} {Explain yourself!
  leveraging language models for commonsense reasoning}.
\newblock In \emph{Proceedings of the 57th Annual Meeting of the Association
  for Computational Linguistics}, pages 4932--4942, Florence, Italy.
  Association for Computational Linguistics.

\bibitem[{Ross et~al.(2021)Ross, Marasovi{\'c}, and
  Peters}]{ross-etal-2021-explaining}
Alexis Ross, Ana Marasovi{\'c}, and Matthew Peters. 2021.
\newblock \href {https://doi.org/10.18653/v1/2021.findings-acl.336} {Explaining
  {NLP} models via minimal contrastive editing ({M}i{CE})}.
\newblock In \emph{Findings of the Association for Computational Linguistics:
  ACL-IJCNLP 2021}, pages 3840--3852, Online. Association for Computational
  Linguistics.

\bibitem[{Slack et~al.(2021)Slack, Hilgard, Lakkaraju, and
  Singh}]{slack2021counterfactual}
Dylan Slack, Anna Hilgard, Himabindu Lakkaraju, and Sameer Singh. 2021.
\newblock Counterfactual explanations can be manipulated.
\newblock \emph{Advances in Neural Information Processing Systems}, 34:62--75.

\bibitem[{Slack et~al.(2020)Slack, Hilgard, Jia, Singh, and
  Lakkaraju}]{10.1145/3375627.3375830}
Dylan Slack, Sophie Hilgard, Emily Jia, Sameer Singh, and Himabindu Lakkaraju.
  2020.
\newblock \href {https://doi.org/10.1145/3375627.3375830} {Fooling lime and
  shap: Adversarial attacks on post hoc explanation methods}.
\newblock In \emph{Proceedings of the AAAI/ACM Conference on AI, Ethics, and
  Society}, AIES '20, page 180–186, New York, NY, USA. Association for
  Computing Machinery.

\bibitem[{Sun et~al.(2022)Sun, Swayamdipta, May, and
  Ma}]{sun-etal-2022-investigating}
Jiao Sun, Swabha Swayamdipta, Jonathan May, and Xuezhe Ma. 2022.
\newblock \href {https://aclanthology.org/2022.findings-emnlp.432}
  {Investigating the benefits of free-form rationales}.
\newblock In \emph{Findings of the Association for Computational Linguistics:
  EMNLP 2022}, pages 5867--5882, Abu Dhabi, United Arab Emirates. Association
  for Computational Linguistics.

\bibitem[{Wang et~al.(2020)Wang, Liang, Jin, Wang, Zhu, and
  Zhang}]{wang-etal-2020-semeval}
Cunxiang Wang, Shuailong Liang, Yili Jin, Yilong Wang, Xiaodan Zhu, and Yue
  Zhang. 2020.
\newblock \href {https://doi.org/10.18653/v1/2020.semeval-1.39}
  {{S}em{E}val-2020 task 4: Commonsense validation and explanation}.
\newblock In \emph{Proceedings of the Fourteenth Workshop on Semantic
  Evaluation}, pages 307--321, Barcelona (online). International Committee for
  Computational Linguistics.

\bibitem[{Wiegreffe and Marasovic(2021)}]{wiegreffe2021teach}
Sarah Wiegreffe and Ana Marasovic. 2021.
\newblock \href {https://openreview.net/forum?id=ogNcxJn32BZ} {{Teach Me to
  Explain: A Review of Datasets for Explainable Natural Language Processing}}.
\newblock In \emph{{Thirty-fifth Conference on Neural Information Processing
  Systems Datasets and Benchmarks Track (Round 1)}}.

\bibitem[{Wiegreffe et~al.(2021)Wiegreffe, Marasovi{\'c}, and
  Smith}]{wiegreffe-etal-2021-measuring}
Sarah Wiegreffe, Ana Marasovi{\'c}, and Noah~A. Smith. 2021.
\newblock \href {https://aclanthology.org/2021.emnlp-main.804} {{M}easuring
  association between labels and free-text rationales}.
\newblock In \emph{Proceedings of the 2021 Conference on Empirical Methods in
  Natural Language Processing}, pages 10266--10284, Online and Punta Cana,
  Dominican Republic. Association for Computational Linguistics.

\bibitem[{Wu et~al.(2021)Wu, Ribeiro, Heer, and Weld}]{wu-etal-2021-polyjuice}
Tongshuang Wu, Marco~Tulio Ribeiro, Jeffrey Heer, and Daniel Weld. 2021.
\newblock \href {https://doi.org/10.18653/v1/2021.acl-long.523} {Polyjuice:
  Generating counterfactuals for explaining, evaluating, and improving models}.
\newblock In \emph{Proceedings of the 59th Annual Meeting of the Association
  for Computational Linguistics and the 11th International Joint Conference on
  Natural Language Processing (Volume 1: Long Papers)}, pages 6707--6723,
  Online. Association for Computational Linguistics.

\bibitem[{Yin et~al.(2022)Yin, Shi, Hsieh, and
  Chang}]{yin-etal-2022-sensitivity}
Fan Yin, Zhouxing Shi, Cho-Jui Hsieh, and Kai-Wei Chang. 2022.
\newblock \href {https://doi.org/10.18653/v1/2022.acl-long.188} {{On the
  Sensitivity and Stability of Model Interpretations in {NLP}}}.
\newblock In \emph{Proceedings of the 60th Annual Meeting of the Association
  for Computational Linguistics (Volume 1: Long Papers)}, pages 2631--2647,
  Dublin, Ireland. Association for Computational Linguistics.

\bibitem[{Yordanov et~al.(2022)Yordanov, Kocijan, Lukasiewicz, and
  Camburu}]{oodNLEs}
Yordan Yordanov, Vid Kocijan, Thomas Lukasiewicz, and Oana-Maria Camburu. 2022.
\newblock {Few-Shot Out-of-Domain Transfer of Natural Language Explanations}.
\newblock In \emph{Proceedings of the Findings of the Conference on Empirical
  Methods in Natural Language Processing (EMNLP)}.

\bibitem[{Yu et~al.(2019)Yu, Chang, Zhang, and
  Jaakkola}]{yu-etal-2019-rethinking}
Mo~Yu, Shiyu Chang, Yang Zhang, and Tommi Jaakkola. 2019.
\newblock \href {https://doi.org/10.18653/v1/D19-1420} {Rethinking cooperative
  rationalization: Introspective extraction and complement control}.
\newblock In \emph{Proceedings of the 2019 Conference on Empirical Methods in
  Natural Language Processing and the 9th International Joint Conference on
  Natural Language Processing (EMNLP-IJCNLP)}, pages 4094--4103, Hong Kong,
  China. Association for Computational Linguistics.

\end{thebibliography}
\bibliographystyle{acl_natbib}


\clearpage
\appendix
\onecolumn
\section{More Examples of Unfaithful NLEs}

\begin{table*}[h!]
\fontsize{10}{8}\selectfont
\centering
\resizebox{\textwidth}{!}{
\begin{tabular}{@{}>{\raggedright\arraybackslash}p{0.08\linewidth}p{0.60\linewidth}p{0.60\linewidth}@{}}
\toprule
\textbf{Dataset} & \textbf{Original Instance} & \textbf{Instance After Test Intervention}\\
\midrule
CoS-E & \textit{Question:} What happens when spending money without paying someone back? \newline \textit{Choices:} poverty, debt, and bankruptcy \newline \textit{Prediction:} debt \newline \textit{NLE:} debt is the only option that is not something that can be paid back. & \textit{\xarrow Question:} What happens when {\color{myblue} increasingly} spending money without paying someone back? \newline \textit{ Choices :} poverty, debt, and bankruptcy. \newline \textit{Prediction:} bankruptcy \newline \xmark  \textit{NLE:} \ul{bankruptcy is the only option that is a result of spending money.} \newline \color{myred}{\textit{Unfaithfulness cause:} inserted word \textbf{`increasingly'} $\notin$ NLE} but changed the prediction.\\

ComVE & \textit{Sent 1:} Everyone hates paying taxes \newline \textit{Sent 2:} Nobody hates paying taxes \newline \textit{Prediction:} first sentence \newline \textit{NLE:} Paying taxes is a good thing & \textit{Sent 1:} Everyone hates paying taxes \newline \textit{\xarrow Sent 2} Nobody {\color{myblue} ardently} hates paying taxes \newline \textit{Prediction:} second sentence \newline \xmark \textit{NLE:} \ul{Paying taxes is a good thing} \newline \color{myred}{\textit{Unfaithfulness cause:} inserted word \textbf{`ardently'} $\notin$ NLE} but changed the prediction. \\

e-SNLI & \textit{Premise:} A man wearing glasses and a ragged costume is playing a Jaguar electric guitar and singing with the accompaniment of a drummer. \newline \textit{Hypothesis:} A man with glasses and a disheveled outfit is playing a guitar and singing along with a drummer. \newline \textit{Prediction:} entailment \newline \textit{NLE:} A ragged costume is a disheveled outfit. & \textit{Premise:} A man wearing glasses and a ragged costume is playing a Jaguar electric guitar and singing with the accompaniment of a drummer. \newline \textit{\xarrow Hypothesis:} A man with glasses and a disheveled outfit is playing a guitar and singing along with a {\color{myblue} semi-formal} drummer. \newline \textit{Prediction:} neutral \newline \xmark \textit{NLE:} \ul{Not all ragged costumes are disheveled.} \newline \color{myred}{\textit{Unfaithfulness cause:} inserted word \textbf{`semi-formal'} $\notin$ NLE} but changed the prediction.  \\

\bottomrule
\end{tabular}}
\caption{Examples of unfaithful explanations detected with \textbf{random insertion baseline}. (see \S\ref{sec:method}). The examples are selected for the MT-RA models for all three datasets. We apply the tests on an original instance (second column), which results in a new instance (third column). The parts of the input changed by the test are marked with \xarrow, and the intervention made by the test is in {\color{myblue} blue}. \xmark marks an NLE or a prediction that does not match the expectation, thus pointing to the \ul{underlined} NLE being unfaithful. 
} 
\label{tab:random-examples}
\end{table*}

\begin{table*}[h!]
\fontsize{10}{8}\selectfont
\centering
\resizebox{1\textwidth}{!}{
\begin{tabular}{@{}>{\raggedright\arraybackslash}p{0.08\linewidth}p{0.60\linewidth}p{0.60\linewidth}@{}}
\toprule
\textbf{Dataset} & \textbf{Original Instance} & \textbf{Instance After Test Intervention}\\
\midrule
CoS-E & \textit{Question:} Where can books be read? \newline \textit{Choices:} shelf, table, and backpack \newline \textit{Prediction:} table \newline \textit{NLE:} books are usually read on a table. & \textit{\xarrow Question:} Where {\color{myblue} outside} can books be read? \newline \textit{ Choices :} shelf, table, and backpack. \newline \textit{Prediction:} backpack \newline \xmark  \textit{NLE:} \ul{books are usually stored in backpacks.} \newline \color{myred}{\textit{Unfaithfulness cause:} inserted word \textbf{`outside'} $\notin$ NLE} but changed the prediction.\\

ComVE & \textit{Sent 1:} When people are hungry they drink water and do not eat food. \newline \textit{Sent 2:} People eat food when they are hungry. \newline \textit{Prediction:} first sentence \newline \textit{NLE:} Water is not a food and cannot satisfy people's hunger. & \textit{Sent 1:} When people are hungry they drink water and do not eat food. \newline \textit{\xarrow Sent 2} People eat food {\color{myblue} so many times} when they are hungry. \newline \textit{Prediction:} second sentence \newline \xmark \textit{NLE:} \ul{Eating food is not a good way to get rid of hunger.} \newline \color{myred}{\textit{Unfaithfulness cause:} inserted words \textbf{`so many times'} $\notin$ NLE} but changed the prediction. \\

e-SNLI & \textit{Premise:} Two women having drinks at the bar. \newline \textit{Hypothesis:} Three women are at a bar. \newline \textit{Prediction:} contradiction \newline \textit{NLE:} Two women are not three women. & \textit{Premise:} Two women having drinks at the bar. \newline \textit{\xarrow Hypothesis:} Three women are {\color{myblue}together} at a bar. \newline \textit{Prediction:} entailment \newline \xmark \textit{NLE:} \ul{Two women are three women.} \newline \color{myred}{\textit{Unfaithfulness cause:} inserted word \textbf{`together'} $\notin$ NLE} but changed the prediction.  \\

\bottomrule
\end{tabular}}
\caption{Examples of unfaithful explanations detected with \textbf{counterfactual editor}. (see \S\ref{sec:method}). The examples are selected for the MT-RA models for all three datasets. We apply the tests on an original instance (second column), which results in a new instance (third column). The parts of the input changed by the test are marked with \xarrow, and the intervention made by the test is in {\color{myblue} blue}. \xmark marks an NLE or a prediction that does not match the expectation, thus pointing to the \ul{underlined} NLE being unfaithful. 
} 
\label{tab:editor-examples}
\end{table*}



\begin{table*}[t]
\fontsize{10}{8}\selectfont
\centering
\resizebox{1\textwidth}{!}{
\begin{tabular}{@{}>{\raggedright\arraybackslash}p{0.08\linewidth}p{0.60\linewidth}p{0.60\linewidth}@{}}
\toprule
\textbf{Dataset, Model} & \textbf{Original Instance} & \textbf{Instance After Test Intervention} \\
\midrule
ComVE, ST-RE & \textit{Sent 1:} Crack addicts are addicted to chocolate milk. \newline \textit{Sent 2:} Crack addicts are addicted to crack. \newline \textit{Prediction:} first sentence  \newline \textit{Explanation:} \ul{Chocolate milk is not addictive.} &  \vspace*{-1.5ex}\xarrow \textit{Sent 1:} {\color{myblue} Crack addicts are addicted to chocolate milk.} \newline \xarrow \textit{Sent 2:} {\color{myblue} Chocolate milk is not addictive.} \newline \xmark \textit{Prediction:} second sentence \newline \textit{Explanation:} Chocolate milk contains a lot of addictive chemicals. \\

ComVE, ST-RA & \textit{Sent 1:} He visited a doctor to cure his sickness \newline \textit{Sent 2:} He went to a lawyer to cure his sickness \newline \textit{Prediction:} second sentence \newline \textit{Explanation:} \ul{Lawyers do not treat people.} & \vspace*{-1.5ex} \xarrow \textit{Sent 1:} {\color{myblue} Lawyers do not treat people.} \newline \xarrow \textit{Sent 2:} {\color{myblue} He went to a lawyer to cure his sickness} \newline \xmark \textit{Prediction:} first sentence \newline \textit{Explanation:} Lawyers treat people \\

ComVE, MT-RE & \textit{Sent 1:} Giraffes have long necks. \newline \textit{Sent 2:} Monkeys have long necks. \newline \textit{Prediction:} second sentence \newline \textit{Explanation:} \ul{Monkeys have short necks.} & \vspace*{-1.5ex} \xarrow \textit{Sent 1:} {\color{myblue} Monkeys have short necks. }\newline \xarrow \textit{Sent 2:} {\color{myblue} Monkeys have long necks. }\newline \xmark \textit{Prediction:} first sentence \newline \textit{Explanation:} Monkeys have long necks. \\

ComVE, MT-RA & \textit{Sent 1:} My knee was scrapped and I put ointment on it. \newline \textit{Sent 2:} My knee was scrapped and I put dirt on it. \newline \textit{Prediction:} first sentence \newline \textit{Explanation:} \ul{Ointment is not used to scrape a knee.} & \vspace*{-1.5ex} \xarrow \textit{Sent 1:} {\color{myblue} My knee was scrapped and I put ointment on it.} \newline \xarrow \textit{Sent 2:} {\color{myblue} Ointment is not used to scrape a knee.} \newline \xmark \textit{Prediction:} second sentence \newline Explanation: Ointment is used to scrape a knee. \\

e-SNLI, ST-RE & \textit{Premise:} People are riding bicycles in the street, and they are all wearing helmets. \newline \textit{Hypothesis:}  A group of friends are grabbing their bikes, getting ready for the morning bike ride. \newline \textit{Prediction:} contradiction \newline \textit{Explanation:} \ul{Just because people are riding bicycles does not mean they are friends.} & \vspace*{-1.5ex} \xarrow \textit{Premise:} {\color{myblue} People are riding bicycles.} \newline \xarrow \textit{Hypothesis:} {\color{myblue} They are friends.} \newline \xmark \textit{Prediction:} neutral \newline \textit{Explanation:} People riding bicycles are not necessarily friends. \\

e-SNLI, ST-RA & \textit{Premise:} A woman is walking her dog and using her cellphone. \newline \textit{Hypothesis:} The woman is playing a game on her cellphone. \newline \textit{Prediction:} neutral \newline \textit{Explanation:} \ul{Just because a woman is using her cellphone does not mean she is playing a game.} & \vspace*{-1.5ex} \xarrow \textit{Premise:} {\color{myblue} A woman is using her cellphone.} \newline \xarrow \textit{Hypothesis:} {\color{myblue} She is playing a game.} \newline \xmark \textit{Prediction:} contradiction \newline \textit{Explanation:} The woman can not be using her cellphone and playing a game at the same time. \\

e-SNLI, MT-RE & \textit{Premise:} A person wearing a red shirt is falling off a white surfboard.\newline \textit{Hypothesis:} A surfer is being flung in to the waves after hiting a large crest. \newline \textit{Prediction:} neutral \newline \textit{Explanation:} \ul{Just because a man is confused doesn't mean he doesn't know where he is.} &\vspace*{-1.5ex}  \xarrow \textit{Premise:} {\color{myblue} A person falling off a surfboard.} \newline \xarrow \textit{Hypothesis:} {\color{myblue} The surfer is being flung in to the waves.} \newline \xmark \textit{Prediction:} entailment \newline \textit{Explanation:} A person falling off a surfboard is being flung in to the waves. \\

e-SNLI, MT-RA & \textit{Premise:} Many people standing outside of a place talking to each other in front of a building that has a sign that says `HI-POINTE.' \newline \textit{Hypothesis:} The people are having a chat before going into the work building. \newline \textit{Prediction:} neutral \newline \textit{Explanation:} \ul{Just because people are talking does not mean they are having a chat.} & \vspace*{-1.5ex} \xarrow \textit{Premise:} {\color{myblue} People are talking.} \newline \xarrow \textit{Hypothesis:}{\color{myblue} They are having a chat.} \newline \xmark \textit{Prediction:} entailment \newline \textit{Explanation:} People are talking is a rephrasing of they are having a chat. \\

\bottomrule
\end{tabular}}
\caption{Examples of unfaithful explanations detected with \textbf{the Input Reconstruction Test}. (see \S\ref{sec:method}). We apply the test on an original instance (second column), which results in a new instance (third column). The parts of the input changed by the test are marked with \xarrow, and the intervention made by the test is in {\color{myblue} blue}. \xmark marks an NLE or a prediction that does not match the expectation, thus pointing to the \ul{underlined} NLE being unfaithful. \color{myred}{ The unfaithfulness cause for the instances is that the reasons in the NLE for the original instance lead to a different prediction.}}
\label{tab:cosistency:examples}
\end{table*}


\subsection{Model Performance}
\begin{table}[t]
\centering
\begin{tabular}{lrr}
\toprule
\textbf{Model} &  \textbf{Acc$\uparrow$} & \textbf{BLEU$\uparrow$}\\
\midrule
\multicolumn{3}{c}{\textbf{SNLI}} \\
MT-Re & 88.24 & 20.01 \\
ST-Re & 87.68  & 19.67\\
MT-Ra & 88.10 & 20.67\\
ST-Ra & 87.63 & 20.59 \\
\multicolumn{3}{c}{\textbf{CoS-E}} \\
MT-Re & 65.79 & 5.75 \\
ST-Re & 66.11 & 6.66 \\
MT-Ra & 66.95 & 5.55 \\
ST-Ra & 67.79 & 7.85 \\
\multicolumn{3}{c}{\textbf{ComVE}} \\
MT-Re & 85.70 & 7.53 \\
ST-Re & 84.40  & 6.68\\
MT-Ra & 86.40  & 7.03 \\
ST-Ra & 86.40  & 7.21\\
\bottomrule
\end{tabular}
\caption{Performance of the models described in Eq \ref{eq:setup}. Acc denotes the prediction performance of the model on the corresponding task. BLEU denotes the BLEU score of the generated explanation compared to the gold human ones.}
\label{tab:performance}
\end{table}

\end{document}